\documentclass[conference]{IEEEtran}
\IEEEoverridecommandlockouts
\usepackage{url}
\usepackage{amsmath,amssymb,amsfonts}
\usepackage{algorithmic}
\usepackage{graphicx}
\usepackage{booktabs}
\usepackage{textcomp}
\usepackage{xcolor}
\usepackage[nolist,nohyperlinks]{acronym}
\acrodef{O/O}{owner/operator}
\acrodefplural{O/O}[O/Os]{owners/operators}
\acrodef{SSN}{Space Surveillance Network}
\acrodef{CDM}{Conjunction Data Message}
\acrodef{TCA}{Time of Closest Approach}
\acrodef{PP}{Poisson Process}
\acrodef{MAE}{Mean Absolute Error}
\acrodef{MSE}{Mean Squared}
\acrodef{RMSE}{Root Mean Squared Error}

\def\BibTeX{{\rm B\kern-.05em{\sc i\kern-.025em b}\kern-.08em
    T\kern-.1667em\lower.7ex\hbox{E}\kern-.125emX}}
\begin{document}

\title{Conjunction Data Messages for Space Collision Avoidance behave as a Poisson Process\\
%
}

\author{\IEEEauthorblockN{1\textsuperscript{st} Francisco M. Caldas}
\IEEEauthorblockA{\textit{NOVA LINCS} \\
\textit{NOVA School of Science and Technology}\\
Caparica, Portugal \\
f.caldas@campus.fct.unl.pt}
\and
\IEEEauthorblockN{2\textsuperscript{nd} Cláudia Soares}
\IEEEauthorblockA{\textit{NOVA LINCS} \\
\textit{NOVA School fo Science and Technology}\\
Caparica, Portugal \\
 claudia.soares@fct.unl.pt}
\and
\IEEEauthorblockN{3\textsuperscript{rd} Cláudia Nunes}
\IEEEauthorblockA{\textit{CEMAT} \\
\textit{Instituto Superior Técnico}\\
Lisboa, Portugal  \\
cnunes@math.tecnico.ulisboa.pt}
\and
\IEEEauthorblockN{4\textsuperscript{th} Marta Guimarães}
\IEEEauthorblockA{
\textit{Neuraspace}\\
Lisboa, Portugal \\
marta.guimaraes@neuraspace.com}

}

\maketitle

\begin{abstract}
  Space debris is a major problem in space exploration. International bodies continuously monitor a large database of orbiting objects and emit warnings in the form of conjunction data messages. An important question for satellite operators is to estimate when fresh information will arrive so that they can react timely but sparingly with satellite maneuvers. We propose a statistical learning model of the message arrival process, allowing us to answer two important questions: (1) Will there be any new message in the next specified time interval? (2) When exactly and with what uncertainty will the next message arrive? The average prediction error for question (2) of our Bayesian Poisson process model is smaller than the baseline in more than 4 hours in a test set of 50k close encounter events. 
\end{abstract}


\section{Introduction}

Since the early 1960s, the space debris population has extensively increased~\cite{radtke2017interactions}. It is estimated that more than 36,000 objects larger than 10 centimetres, and millions of small er pieces, exist in Earth's orbit \cite{esa_2021}. Collisions with debris give rise to more debris, leading to more collisions in a chain reaction known as Kessler syndrome~\cite{krag20171}. To avoid catastrophic failures, satellite \acp{O/O} need to be aware of the collision risk of their assets~\cite{le2018space}. Currently, this monitoring process is done via the global \ac{SSN}.  To assess possible collisions, a physics simulator uses \ac{SSN} observations to propagate the evolution of the state of the objects over time~\cite{horstmann2017investigation,mashiku2019recommended}. Each satellite (also referred to as target) is screened against all the objects of the catalogue in order to detect a conjunction, i.e., a close approach. Whenever a conjunction is detected between the target and the other object (usually called chaser), \ac{SSN} propagated states become accessible and a \ac{CDM} is issued, containing information about the event, such as the \ac{TCA} and the probability of collision. Until the \ac{TCA}, more \acp{CDM} are issued with updated and better information about the conjunction. Roughly in the interval between two and one day prior to \ac{TCA}, the \acp{O/O} must decide whether to perform a collision avoidance manoeuvre, with the available information. Therefore, the \ac{CDM} issued at least two days prior to \ac{TCA} is the only guaranteed information that the \acp{O/O} have and, until new information arrives, the best knowledge available.

Several approaches have been explored to predict the collision risk at \ac{TCA}, using statistics and machine learning~\cite{acciarini2021kessler}, but only a few were developed with the aim of predicting when the next \ac{CDM} is going to be issued. Very recently, \cite{pinto-2020-automate} developed a recurrent neural network architecture to model all \ac{CDM} features, including the time of arrival of future \acp{CDM}. GMV is currently developing an autonomous collision avoidance system that decides if the current information is enough for the \acp{O/O} to decide, or if they should wait for another CDM to have more information .
However, the techniques used are not publicly available. In this work, we propose a novel statistical learning solution for the problem of modeling and predicting the arrival of \acp{CDM} based on a homogeneous \ac{PP} model, with Bayesian estimation. We note that standard machine learning and statistical learning solutions are data-hungry and cannot model our problem, suffering from a special type of data scarcity. Further, as our application is high-stakes, we require confidence or credibility information added to a point estimate. 

\subsection{Background} The present work formalizes the problem of modelling and predicting the arrival of a new \ac{CDM} with a probabilistic generative model. The present formulation has the advantage of providing a full description of the problem, stating clearly the required assumptions. Our proposed model shows high accuracy, decreasing error in more than 4 hours, when compared with the baseline, for predicting the next event.
\begin{figure}[t]
    \centering
    \includegraphics[scale=0.8]{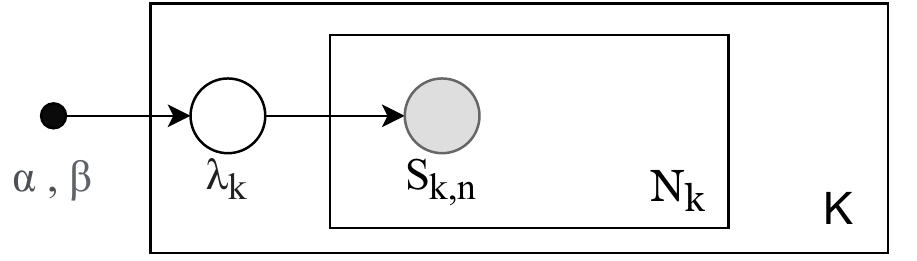}
    \caption{Graphical model representation of the proposed model. The boxes are plates representing replication of the structure. The outer plate represents the dataset of $K$ events, and the inner one the repeated prediction of $N_k$ occurrences in each event.}
    \label{fig:graph}
\end{figure}
We leverage on a data model depicted in Figure~\ref{fig:graph}. Probabilistic Graphical Models~\cite{koller2009probabilistic,bishop2006pattern} is a probabilistic framework that allows for highly expressive models, while keeping computational complexity to a minimum. A few statistical learning developments like the Latent Dirichlet Allocation for topic modelling~\cite{blei2003latent}, are now the basis for deep learning-powered probabilistic models like the Variational Autoencoder~\cite{kingma2019introduction} and Normalizing flows~\cite{kobyzev2020normalizing}, and as a means of creating modular and interpretable machine learning models~\cite{murdoch2019definitions}. We leverage on the homogeneous Poisson Point Process~\cite{kulkarni2016modeling}, a counting process of messages occurring over continuous time. Thus, we assume that the inter-CDM times are independent and identically distributed random variables with exponential distribution with a rate~$\lambda$. The homogeneity property comes from an event rate constant in time. Phenomena like radioactive decay of atoms and website views are well modeled by this process~\cite{ross2014introduction}. The homogeneity is, nevertheless, a limitation that will be addressed in future work, as there might be some temporal trend in the data. Not considering it, though, delivers a fast to compute solution that can outperform a standard baseline.
\subsection{Contributions}
\begin{itemize}
    \item Formalizing an arrival problem within a probabilistic graphical model framework
    \item Few-shot learning using a Bayesian Approach
    \item Model testing with a real world problem, surpassing the Baseline
    \item analysis of the limitations of the homogeneous Poisson Process
\end{itemize}

\section{Poisson Point Process}

A homogeneous \acf{PP} can be seen as a counting process of events over continuous time. A process is said to be Poisson when the inter-event times are independent and identically distributed random variables with exponential distribution and parameter $\lambda$. If $\lambda$ is constant over the continuous time interval, then the \ac{PP} is homogeneous \cite{kulkarni2016modeling}.

A Poisson Process is a specific type of Renewal Process, when the interarrival times have an exponential distribution. A renewal process may have any independent and identically distributed (i.i.d.) inter-event times with finite mean. 

The Poisson process related with the \acp{CDM} is defined as follows:

\begin{enumerate}

    \item \{${X_i} : i \in \mathbb{N} \}$ be a sequence of random variables, where $X_i$ denotes the inter-CDM time, with $X_i \sim$ Exponential($\lambda$);
    \item Then if we define $S_n  = \sum_{i=1}^n X_i$, it follows that $S_n$ is the time of occurrence of the $n_{th}$ \ac{CDM};
    
    \item Finally, $N(t) = \text{max}\{n \in \mathbb{N}_0 :S_n \leq t\},\text{for}~t \geq 0$ is the number of \acp{CDM} issued in the interval (0,t].
    
\end{enumerate}

The counting process $N(t)$ is therefore a \ac{PP} with rate $\lambda$, i.e., $$ \{N(t):t \geq 0\} \sim PP(\lambda)$$

There are two very important characteristics in a Poisson process. For any interval $[a,b]$ let $N([a,b])$ denote the number of \acp{CDM} at that interval. There, from the fact that $\{N(t):t \geq 0\}$ is a \ac{PP}, it follows that $N([a+s,b+s])$ is Poisson distributed, with parameter equal to $\lambda (b-a)$, for all $s,a,b \geq 0$ (stationary increments). Furthermore, given two disjoint time intervals, $[a,b]$ and $[c,d]$, $N([a,b])$ and $N([c,d])$ are independent random variables (independent increments).

Finally, we postulate that each event has a specific unknown arrival rate $(\lambda_k)$, and the determination of that parameter is the objective of the graphical model.

\section{Problem Formulation}

Our generative probabilistic model of the CDM arrival process will allow answering some questions as: (1) what is the probability that, during the decision phase, new information will be received (2) what is the best estimate for the next CDM arrival time, and the uncertainty of this prediction.

\bigskip

Following the literature, we define the stochastic quantities, with $k$ as the ID of the object:
\begin{itemize}
    \item $X_{n+1}^k$ is time (in days) between the  $n^{th}$ and the $n+1^th$ CDM, where $X_{n+1}^k \sim Exp(\lambda_k)$, i.e., $X_{n+1}$ has an exponential distribution of rate $\lambda_k$;
    \item $N^k(t) $ is the number of CDMs received in the interval (0,t], in days, where  $N^k(t) \sim \text{Poisson}(\lambda_k t)$, and $\lambda_k$ is the rate of CDMs issued per day;
    \item $S_n^k  = \sum_{i=1}^n X_i^k$ is the time of occurrence of the $n^{th}$ CDM, which, in view of the definition of $X_n^k$ is such that:
    $$S_n^k \sim Gamma(n,\lambda_k)$$.
\end{itemize}

\bigskip
\noindent \textbf{Probability of Interest}:
We are interested in the probability of receiving one or more CDMs in the time interval from the last available observation $s_n^k$, until a constant security threshold $(t_{sec})$. In this application, we consider 1.3 days prior to TCA, and the interval will be $(s_n^k,t_{sec}]$.
\begin{align}
    P(N^k(t_{sec}-s_n^k) \geq 1) &= 1 - P(N^k(t_{sec}-s_n^k) =0)   \label{eq:prop} \\ 
 & = 1- e^{-\lambda_k(t_{sec}-s_n^k)}\nonumber
   \label{eq:prop_poisson}
\end{align}

\noindent \textbf{Point Prediction}: Furthermore, as:
\begin{align*}
  &S^k_{n+1} = S^k_n + X^k_n \qquad \qquad \qquad 
  X^k_n \sim Exp(\lambda_k)  \\
  &\text{it follows that:} \\
& (S^k_{n+1} -S^k_n | s^k_{n}) \sim Exp( \lambda_k) \\
\end{align*}

and therefore we may estimate the time until the next CDM to be issued by $\frac{1}{\hat{\lambda^k}}$, the estimate of the expected value of the inter-CDM time for event $k$.

\subsection{Baseline}
The conjunction events are composed of a series of real \acp{CDM} for a set of satellites during a 4 year period. The training data set has around 50,000 independent conjunctions events, with each event having at least 2 messages and 15 messages in average. The complete dataset and variables are described in \cite{uriot2020spacecraft}, however, in this case, the only required information are the times between message arrivals for each event. 
We assume the baseline model to be the simplest one, where we assume that the next inter-CDM time is equal to the last one, i.e.: 

$$ X_{n+1}^k = x_{n}^k $$ for each event k. This baseline implies a Markovian property on the variable of interest.

In addition, we will also consider another model --- hereby called by Classical --- which is nothing more than estimating $\lambda_k$ as in the classical frequentist estimation. Because, for most events, the number of inter-CDM times is very small, we expect this estimator process will not be as robust and will provide biased results. Therefore, we need other models, besides the Baseline and the Classical PP.


\section{Bayesian Approach}

 To have a statistically significant estimation of $\lambda_k$, and in order to use information from the other events, we adopt the Bayesian model presented in Figure~\ref{fig:graph}. More specifically, we use the extensive amount of events in the dataset with an empirical Bayes framework to get an informative prior, and then we update that distribution with the specific inter-CDMs times of each event. We end up with a posterior distribution for $\lambda_k$, of which we can extract a Maximum a Posteriori (MAP) estimate, $\hat{\lambda}_k$ \cite{ruggeri}. This method is basically described in the following steps:

The posterior $\lambda$ distribution is defined as:
\begin{enumerate}
    \item the prior distribution for $\lambda_k$ is $$\lambda_k \sim \text{Gamma}(\alpha,\beta)$$
    where $\alpha$ and $\beta$ are hyperparameters estimated using data for all the events in the train data.
    \item The likelihood function, for each event $k$, based on the information $(X_1^k,X_2^k,...,X_n^k)$ is given by:
    \begin{equation}
        l(\lambda_k|X_1^k,X_2^k,...,X_n^k) = \lambda_k^n e^{-\lambda_k \sum^n_{i=1}X_i^k}
    \end{equation}
    where $(X_1^k,X_2^k,...,X_n^k)$ is the information regarding the inter-CDM times for the $k^{th}$ event
    \item Then, it follows from straightforward manipulations, that the posterior distribution of $\lambda_k$ is \cite{kulkarni2016modeling}:
    \begin{align}
    f(\lambda_k| n, T ) \propto \lambda_k^n e^{-\lambda_k T}  \lambda_k^{\alpha-1}e^{-\beta \lambda_k} \nonumber \\
    f(\lambda_k| n, T ) \propto \lambda_k^{\alpha + n -1} e^{-\lambda_k(\beta+T)} \nonumber\\
    \lambda_k| n,T \sim \text{Gamma}( \alpha + n, \beta + T^k).
    \label{eq:po}
    \end{align}
    With $T^k=\sum^n_{i=1}X_i^k$.
\end{enumerate}

Note that the posterior in \eqref{eq:po} can be derived either from the set of inter-event times or through the number of CDMs received in a given interval $N^k(t)$.

\begin{figure}[htb!]
    \centering   \includegraphics[width=0.95\columnwidth]{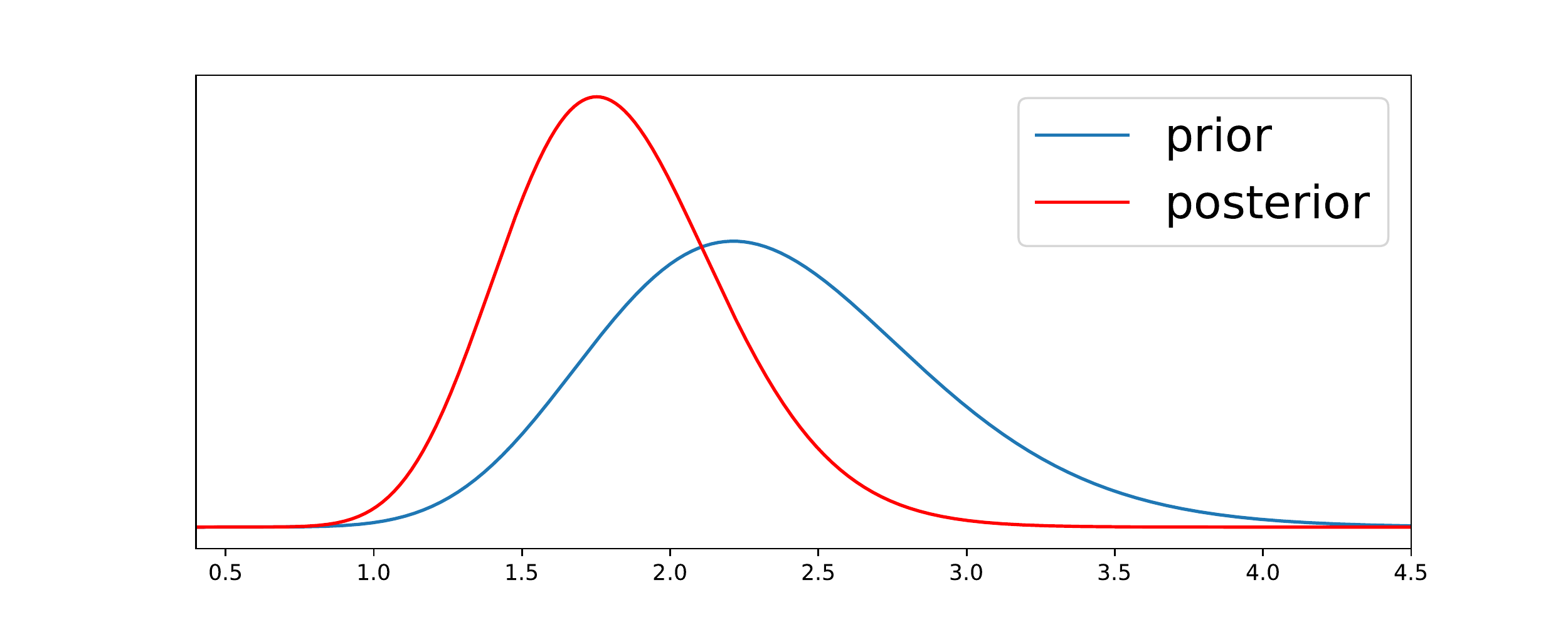}
    \caption[Prior and Posterior Dis. $\lambda$]{Prior distribution of $\lambda$, and a posterior distribution for a single event in the train data.}
    \label{fig:prior_posterior}
\end{figure}

In Figure \ref{fig:prior_posterior}, we observe the initial trained prior, and an example of a posterior distribution, for a particular event. We can see that while informative, the prior has enough variance that with few updates there is a shift in the distribution of the posterior rate of arrival ($\lambda_k$). We can also see that the posterior is narrower, meaning that the credible interval of the posterior is smaller.

\section{Results}

Next, we present the results obtained with the Bayesian, Classical and Baseline models. For the computation of the estimations errors, we have used half of the data as training data and the other half as an unbiased test set.
In Table \ref{tab:plain}, we show the prediction errors (according to the measures: \acf{MAE}, \acf{MSE} and \acf{RMSE}) for the three models.

\begin{table}[htb]
\caption[Poisson Process Results]{\acs{MAE}, \acs{MSE} and \acs{RMSE} for next CDM time prediction, in days.}
\centering
\begin{tabular}{lccc}
\hline
  & MAE & MSE & RMSE\\
\hline
Bayesian   &  0.15170   &    0.06282  & 0.25064 \\
Classic &  0.17581   &   0.08789  & 0.29646 \\
Baseline & 0.26075 &   0.19105 & 0.43709 \\
\hline
\end{tabular}
\label{tab:plain}
\end{table}

\begin{figure}[htb!]
    \centering
    \includegraphics[width=0.95\columnwidth]{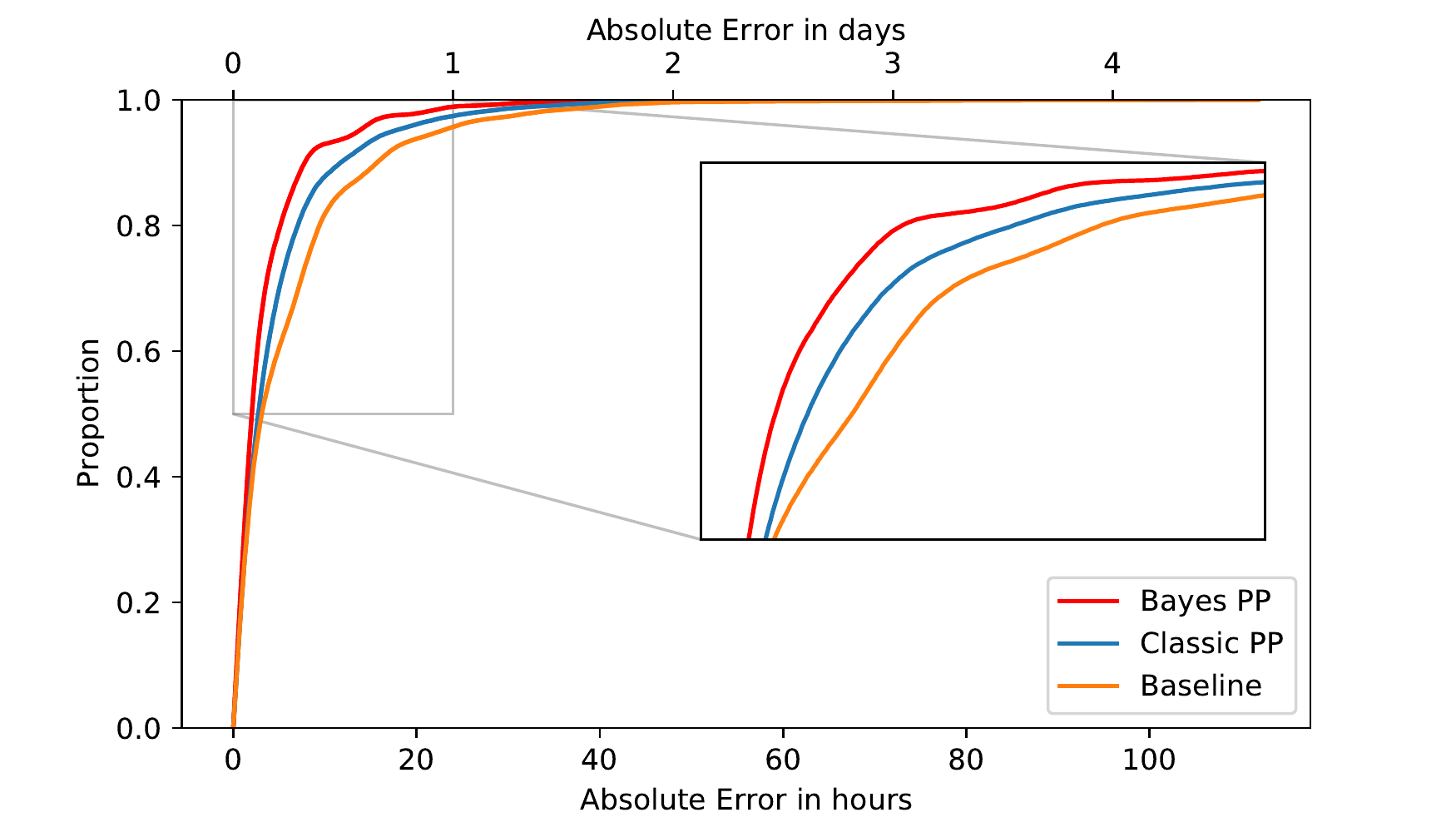}
    \caption[Empirical CDF]{Empirical CDF for each of the models. There is a perceivable difference in the predictive capability of each model. Because the three models use the same formula to predict the expected time of the next CDM, the difference lies in parameter determination, that is, the rate of arrival $\lambda_k$.}
    \label{fig:cdf}
\end{figure}

These results were obtained using an unbiased test data of 50523 independent events, that was not seen during the study and was not used to derive the hyperparameters. The \ac{RMSE} for the proposed Bayesian model is $0.25064$ days, which corresponds to six hours. 
Comparing the Bayesian with the classical estimation of the parameter, we note that we expect better accuracy, as the number of CDMs for each event is small, and thus, the classical approach is more sensitive to extreme values.
As can be seen in Figure~\ref{fig:cdf}, the proposed model outperforms the classic counterpart and the baseline.

\begin{figure}[htb!]
    \centering
    \includegraphics[scale=0.5]{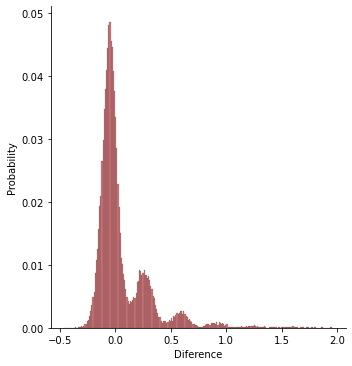}
    \caption[Bayesian PP error distribution]{Error Distribution for the Bayesian Poisson Process. Difference in days between the Predicted Time of Arrival and the Observed Time of Arrival.}
    \label{fig:hist_dif}
\end{figure}

A more detailed look into the performance of the Bayesian model, leads us to an interesting observation. From Figure~\ref{fig:hist_dif}, we conclude that the distribution of the error is not Gaussian.

\noindent This might indicate that our assumptions are not adapted to the data that we are analysing. This means, in particular, that the assumption of exponential distribution for inter-CDM times may not be adequate. Another problem that may cause this behaviour is related with the fact that some \acp{CDM} are not issued. In fact, there might be a probability $p>0$ that a  \ac{CDM} is issued but not received (leading to a filtered Poisson Process), then the apparent observed time of the $n+1$ \ac{CDM}, is actually the time of the $n+2$ \ac{CDM}.

In order to confirm the good performance of our approach, we compare the estimated probability of receiving a CDM in a decision interval with the empirical probability. The estimated probability is the probability obtained using \eqref{eq:prop}. To obtain the empirical probability, the events are grouped by the estimated probability intervals as presented in Table \ref{tab:prop}. By computing ratio of events in each group that actually receive a CDM in that time interval, thus getting an empirical probability.
For example, for events which where determined to have an estimated probability in the interval $(0.803,0.952]$ with equation \eqref{eq:prop}, we analyzed how many of them actually received a CDM in the time interval. 
We note that there is a small positive drift between the estimated probability under the Poisson assumption and the empirical one, meaning that the estimated probability is conservative when compared to reality. We also note that, as it should be, a lower estimated probability will indeed represent a lower empirical probability, because along the groups, the empirical probability increases.

\begin{table}[htb]
\caption[Probability of receiving CDM]{Grouping of events by estimated probability, and the empirical results for each group. The Estimated probability is not over-confident when determining if a CDM is going to be issued during the decision interval.}
\centering

\begin{tabular}{llr}
\toprule
Estimated Prob.  &  Empirical Prob.  & Deviation \\
\midrule
(0.704, 0.753]    & 0.7647  &  0.0117  \\
(0.753, 0.803]   & 0.83992  &    0.03692     \\
(0.803, 0.852]  & 0.90365 &  0.05165     \\
(0.852, 0.901]  & 0.9401  &  0.0391  \\
(0.901, 0.951]  & 0.96740  & 0.0164  \\
(0.951, 1.0]   & 1.0  & 0\\
\bottomrule

\end{tabular}
\label{tab:prop}
\end{table}

\begin{figure}[htb]
    \centering
    \includegraphics[scale=0.68]{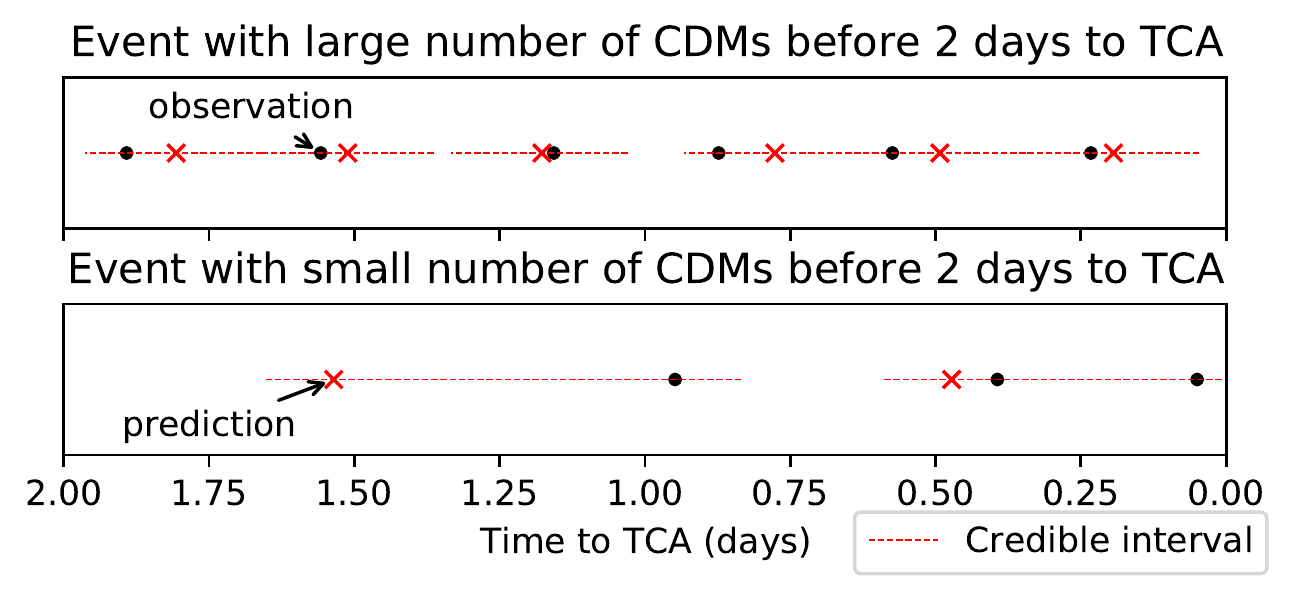}
    \caption[Prediction Example]{Example of the predictive capability of the model for different examples. Top panel uses 15 observations to update the prior. Bottom panel uses 3 observations. }
    \label{fig:examples}
\end{figure}

In Figure \ref{fig:examples} we can see two examples of the temporal prediction for two events in our test data, with 90\% credible intervals. This figure shows that when the number of observations used for the Bayesian estimation is not too low (15, in top panel), the observed and estimated times are close, and more important from a statistical point of view, the observed value is contained within the credibility interval. 
In the lower panel case, the credible intervals have a large range (spanning over almost one day), and the observations and predictions are far apart. Therefore when the number of observations in low, the prediction results are less meaningful.

\section{Conclusion and Future Work}
To conclude this section, we found that this real-life problem of predicting the time of arrival of the next \ac{CDM} can be successfully modeled by a stochastic process, and that by using a Bayesian scheme, we can overcome data scarcity. By modeling this problem as a \ac{PP}, we can estimate the arrival of CDMs during the decision period, which can be used in practice to aid with expert decision process, helping the operator to confidently delay the manoeuvre decision until new information is received. The proposed model is able to predict the time of the next CDM with accuracy exceeding both the baseline and a simpler predictor. The error distribution of the model suggests further improvements to the model, such as the generalization of this model as a renewal process, or the indication that the model should be a filtered Poisson Process.
The model is also slightly conservative when it comes to the empirical probability of receiving a CDM in a given interval, and this might mean that the initial hypothesis of homogeneity might not completely match to the data. While the deviation is small, it is consistent, and it might indicate that in the last two days, there is a bias in the model to predict a smaller number of CDMs. However, for the practical case of a high-stakes application like space awareness, it is always safer to be under-confident than over-confident.

\section*{Acknowledgment}
The authors of this paper would like to thank FCT/PT Space under the PhD grant PRD/BD/153601/2021 and Neuraspace for supporting this research.

%
%
%
\bibliographystyle{IEEEtran} 
\bibliography{IEEEexample}

\end{document}